\documentclass[10pt,twocolumn,letterpaper]{article}

\usepackage{cvpr}
\usepackage{times}
\usepackage{epsfig}
\usepackage{graphicx}
\usepackage{amsmath}
\usepackage{amssymb}

\usepackage{booktabs}
\usepackage{adjustbox}
\usepackage{fixltx2e}
\usepackage{subcaption}
\usepackage[breaklinks=true,bookmarks=false]{hyperref}

\newcommand\blfootnote[1]{%
 \begingroup
 \renewcommand\thefootnote{}\footnote{#1}%
 \addtocounter{footnote}{-1}%
 \endgroup}
\newcommand*\rot{\rotatebox{90}}

\cvprfinalcopy
\def\cvprPaperID{2025} 

\ifcvprfinal\fi
\begin{document}

\title{{\em Shoestring:} Graph-Based Semi-Supervised Classification with Severely Limited Labeled Data}

\author{Wanyu Lin,~Zhaolin Gao,~Baochun Li\\
University of Toronto\\
{\tt\small \{wanyu.lin, zhaolin.gao\}@mail.utoronto.ca}~
{\tt\small bli@ece.toronto.edu}
}

\maketitle
\thispagestyle{empty}

\begin{abstract}

Graph-based semi-supervised learning has been shown to be one of the most effective classification approaches, as it can exploit connectivity patterns between labeled and unlabeled samples to improve learning performance. However, we show that existing techniques perform poorly when labeled data are severely limited. To address the problem of semi-supervised learning in the presence of severely limited labeled samples, we propose a new framework, called {\em Shoestring}\footnote{Code will be made publicly available.}, that incorporates metric learning into the paradigm of graph-based semi-supervised learning. In particular, our base model consists of a graph embedding network, followed by a metric learning network that learns a semantic metric space to represent the semantic similarity between the sparsely labeled and large numbers of unlabeled samples. Then the classification can be performed by clustering the unlabeled samples according to the learned semantic space. We empirically demonstrate Shoestring's superiority over many baselines, including graph convolutional networks, label propagation and their recent label-efficient variations (IGCN and GLP). We show that our framework achieves state-of-the-art performance for node classification in the low-data regime. In addition, we demonstrate the effectiveness of our framework on image classification tasks in the few-shot learning regime, with significant gains on miniImageNet ($2.57\%\sim3.59\%$) and tieredImageNet ($1.05\%\sim2.70\%$).

\end{abstract}

\blfootnote{This research was supported in part by the NSERC Discovery Research Program.}

\section{Introduction}
\label{sec:intro}

The availability of large quantities of labeled samples has made it possible for deep learning to achieve remarkable performance breakthroughs in speech recognition, natural language processing, and computer vision~\cite{he2016deep,simonyan2014very}. However, the reliance on large amounts of labeled samples increases the burden of data collection, making it difficult to apply deep learning to the low-data regime where labeled samples are extremely rare and are difficult to collect. 

With semi-supervised learning (SSL), small amounts of labeled samples are used with a relatively large number of unlabeled samples for classification. Among existing semi-supervised learning models, graph-based methods, such as graph convolutional networks and label propagation, have been demonstrated as one of the most effective approaches for semi-supervised classification, as they are capable of exploiting the connectivity patterns between labeled and unlabeled samples to improve classification performance. Given their advantages, in previous work on few-shot image classification, quick knowledge from a few samples is acquired by considering relationships between instances and representing the data into a graph~\cite{garcia2017few,iscen2019label,jiang2019semi,liu2018learning}. 

Yet, even with such graph-based semi-supervised learning, model learning performance degrades quickly with a diminishing number of labeled samples per class~\cite{li2018deeper}. The performance degradation can be explained as follows. In general, labels work as ``anchors,'' and are used to force the learning models to fit these labeled samples with certain confidence, so that the information extracted from them can be reliably propagated to unlabeled samples. However, when the labeled samples are severely limited, there is a good chance they will exhibit a large testing error even though their training error is small---i.e., overfitting these limited labeled data. Taking graph convolutional networks as an example, they indeed lead to state-of-the-art accuracies on node classification tasks with two convolutional layers in the presence of a sufficient amount of labeled samples. However, when only a few labeled samples are given, it would not be able to effectively propagate the labels to the entire data graph~\cite{li2018deeper}. 

Nevertheless, humans are exceptional learners capable of generalizing their learned knowledge to novel concepts, and capable of learning from very few examples. In this paper, we aim to tackle the problem of graph-based semi-supervised learning where labeled data are {\em severely limited}. There has been a major push in recent research, particularly on the image classification task, towards generalizing deep learning models to learn tasks in a data-efficient way through few-shot learning. 

Among the best-performing methods (e.g., gradient-based~\cite{finn2017model}, metric-learning based~\cite{ren2018meta,snell2017prototypical} and model-based~\cite{mishra2017simple}) for few-shot learning, metric-learning approaches have been demonstrated as one of the simplest and most efficient methods in the few-shot setting. Metric-learning methods aim to optimize the transferable embeddings by learning a distance-based prediction rule over the embeddings. Motivated by this finding, in addition to exploiting the connectivity patterns between labeled and unlabeled samples, we seek to transfer as much knowledge as possible from limited labeled samples to a large number of unlabeled samples in the embedding space. 

The main contribution in our proposed framework, called {\em Shoestring}, is that it is the first to incorporate a metric learning network into the settings of graph-based semi-supervised learning. It is simple yet effective that can be applied to boost the learning performance of typical graph-based semi-supervised learning methods. In essence, our framework is proposed based on the idea that in the low-dimensional semantic space, there exists an embedding in which points cluster around a single prototype representation for each class. More specifically, {\em Shoestring} jointly learns a non-linear mapping of each instance into a semantic space using a graph embedding network, and learns a metric space with a metric learning network to represent the semantic similarity between the labeled and unlabeled samples. Classification, for an embedded unlabeled sample, is then performed by finding its nearest class prototype based on the learned semantic metric.

Highlights of our original contributions are as follows. {\em First}, to verify the effectiveness of our framework, we revisited several graph-based semi-supervised learning models, such as graph convolutional networks, label propagation and their recent label-efficient variations proposed from the perspective of graph filters (IGCN and GLP), and empirically demonstrate the superiority of our framework over these baselines. We show that our framework leads to state-of-the-art node classification performance in the low-data regime, by incorporating these graph learning models as the base models. {\em Second}, we empirically analyze the underlying distance functions used in the metric learning network, such as cosine similarity and squared Euclidean distance. We find that the choice of a similarity metric is critical, as the performance of different metrics varies from different datasets as well as various label rates. {\em Finally}, we demonstrate the effectiveness of {\em Shoestring} on image classification tasks in the few-shot learning regime, and achieve state-of-the-art results on miniImageNet and tieredImageNet.

\section{Problem Setup}
\label{sec:problem}

We consider the task of semi-supervised node classification on graphs. Formally, a graph $\mathcal{G}=\mathcal{(V, A, X)}$ is given with $n=|\mathcal{V}|$ vertices, where $\mathcal{V} = \{v_1, v_2,\cdots, v_n\}$ is the set of vertices, $\mathcal{A}\in \{0,1\}^{n\times n}$ is the adjacency matrix representing the connections, and $\mathcal{X}=\{x_1, x_2,\cdots, x_n\}^T \in \mathcal{R}^{n\times m}$ is the feature matrix of vertices, and $x_i\in \mathcal{R}^m$ is the $m$-dimensional feature vector of vertex $v_i$. 

We follow the standard semi-supervised classification setting, which is commonly employed in various literature~\cite{bengio200611,kipf2016semi}. Given a set of labeled nodes $\mathcal{V}_l\subset \mathcal{V}$, with class labels from $\mathcal{Y}=\{y_1, y_2, y_3,\cdots, y_K\}$ and a set of unlabeled nodes $\mathcal{V}_u \subset \mathcal{V}/ \mathcal{V}_l$, the goal of node classification is to map each node $v\in\mathcal{V}$ to one class in $\mathcal{Y}$. We assume that the data domain is sparsely labeled so that the number of node-label pairs is much smaller than the number of unlabeled nodes, $|\mathcal{V}_l| \ll |\mathcal{V}_u|$. In particular, we are especially interested in cases where $|\mathcal{V}_l|$ is severely limited, e.g., $1$ or $2$ labeled samples per class which may arise in situations where obtaining an unlabeled sample is cheap and easy for novel classes, while labeling the sample is expensive or difficult. Our ultimate goal is to produce an effective classifier for semi-supervised node classification on graphs, for which only very few labeled samples are available.  

\section{Revisiting Graph-based Semi-Supervised Learning}
\label{sec:revisit_ssl}

We do not attempt to provide a comprehensive literature review on graph-based semi-supervised learning. Instead, we selectively provide the baseline methods adopted by top performers on node classification tasks, such as graph convolutional networks and label propagation, either in terms of their simplicity or expressiveness. Furthermore, we think that these methods are of great value, not the least because they lead to state-of-the-art node classification with small numbers of labeled data in the literature and can readily be applied to image classification tasks in the few-shot learning regime~\cite{garcia2017few,liu2018learning}. As prototypical examples, let us consider semi-supervised classification with graph convolutional networks~\cite{kipf2016semi} and label propagation methods~\cite{bengio200611,zhou2004learning,zhu2003semi}.

\textit{Graph convolutional networks:} Graph convolutional neural networks (GCNs) is a generalization of traditional convolutional neural networks to the graph domain. In~\cite{kipf2016semi}, the GCN model applied for semi-supervised classification is a two-layer GCN followed by a softmax classifier on the output features:

\begin{equation}
Z=\mathbf{softmax}(\hat{\mathcal{A}}\mathbf{ReLU}(\hat{\mathcal{A}}\mathcal{X}\Theta^{(0)})\Theta^{(1)})
\end{equation}

where $\tilde{\mathcal{A}} = \mathcal{A}+I$, $\tilde{\mathcal{D}}_{ii}=\sum_{j} \tilde{\mathcal{A}}_{ij}$, $\hat{\mathcal{A}}=\tilde{\mathcal{D}}^{-\frac{1}{2}}\tilde{\mathcal{A}}\tilde{\mathcal{D}}^{-\frac{1}{2}}$, $\mathbf{softmax}(x_i)=\frac{1}{Z}\exp(x_i)$ with $Z=\sum_{i}\exp(x_i)$. The optimization loss function is defined as the cross-entropy error over all labeled samples:
\begin{equation}
\label{eq:gcnloss}
\mathcal{L}_{\mathbf{ce}} = -\sum_{i\in\mathcal{V}_l} \sum_{k=1}^{K} Y_{ik}\ln Z_{ik}
\end{equation}

where $\mathcal{V}_l$ is the set of node indices that have labels, and K is the number of classes/labels.

{\em Label propagation:} Label propagation is a simple and effective principle of using the graph structure to spread labels from labeled samples to the entire data set. Starting with nodes with their known labels, each node starts to propagate its label to its neighbors, and the process is repeated until convergence. Due to its simplicity and effectiveness, there are several variations in the literature~\cite{bengio200611,zhou2004learning,zhu2003semi} and have been widely used in many scientific research fields and numerous industrial applications. An alternative method originating from smoothness considerations yields algorithms based on graph regularization, which naturally leads to a regularization term based on the graph Laplacian. Formally, the objective is to find an embedding matrix $Z$ that agrees with the label matrix $Y$ while being smooth on the graph such that nearby vertices have similar embeddings.

\begin{equation}
Z = \mathbf{arg}\mathbf{min}\{||Z-Y||^2_2+\alpha \mathbf{Tr}(Z^{T}LZ)\}
\end{equation}

where $L=D - \mathcal{A}$ is the graph Laplacian, $D$ is the degree matrix, and $\alpha$ is a parameter controlling the degree of Laplacian regularization. Then a closed-form solution can be obtained by taking the derivative of the objective function and setting it to zero.

\textbf{Analysis} In essence, for semi-supervised learning to work, a certain assumption, called the {\em smoothness assumption}, has to hold. It implies that if two inputs $x_1$, $x_2$ in a high-density region are close, then so should be the corresponding outputs $y_1$, $y_2$. Semi-supervised GCN and label propagation methods have been proved to perform very well on many classification tasks. 

These can be explained as follows. For GCN, graph convolution is a special form of Laplacian smoothing, which computes the new representation of a vertex by averaging over itself and its neighbors. Regarding label propagation, the second term of its objective function is a regularization term motivated by the smoothness assumption. When the number of labeled samples is large enough, both GCN and label propagation can effectively learn the shape of the manifolds near which the data concentrate in the embedding space, leading to superior performance on node classification tasks.

\noindent\textbf{Why do these methods fail?} Graph convolutional networks and label propagation essentially fall into the category of local learning algorithms in semi-supervised learning, relying on a neighborhood graph to approximate manifolds near which the data density is assumed to concentrate. When there are only a few labeled samples, one cannot generalize properly and the model performance degrades very quickly. 

\noindent{\textbf{Graph filtering-based variations of GCN and LP with severely limited labeled samples.}}~\cite{li2019label} aims to address the problem of label efficient semi-supervised learning from the perspective of graph filtering. They proposed a framework that draws graph structure into data features by taking them as signals on the graph and applying a low-pass graph filter to extract data representations for downstream classification tasks. Indeed, it can achieve label efficiency, to some extent, by adjusting the strength of the graph filter. Under this framework, generalized label propagation (GLP) and improved graph convolutional networks (IGCN) were proposed with two types of variations respectively, either relying on the renormalization (RNM) filter or the auto-regressive (AR) filter. 

We evaluated the task of document classification with different semi-supervised learning methods on Cora~\cite{mccallum2000automating} and CiteSeer~\cite{giles1998citeseer} respectively, each of which has one labeled sample per class. The results are shown in Table~\ref{tab:cora}. We observed that with severely limited labeled samples, the performance of graph filtering-based variations are non-significant. More specifically, IGCN performs worse than GCN on CiterSeer, while GLP leads to degraded performance on Cora. In this paper, we are interested to exploit the intrinsic structure of the data to boost classification accuracy with further gains when the number of labeled samples is severely limited. 

\begin{table}[!t]
\caption{Classification accuracy on Cora and CiteSeer, with one labeled sample per class ($\%$). The performance of graph filtering-based variations are insignificant in the severely low-data regime.}
\begin{center}
\begin{small}
\begin{sc}
\setlength{\tabcolsep}{0.25em}
\begin{tabular}{ c |c |c |c| c|c|c}
\toprule
 Dataset& GCN & IGCN\textsubscript{(RNM)} & IGCN\textsubscript{(AR)}& LP  & GLP\textsubscript{(RNM)} & GLP\textsubscript{(AR)}\\ 
\midrule
 Cora&$39.5$ & $41.5$ & $42.3$  & $43.6$& $38.4$ &$37.7 $\\
 CiteSeer&$34.1$&$33.1$&$33.0$&$30.6$&$37.0$&$37.4$\\
\bottomrule
\end{tabular}
\end{sc}
\end{small}	
\end{center}
\label{tab:cora}
\end{table}

\begin{figure*}[!ht]
  \begin{center}
  \includegraphics[scale = 0.23]{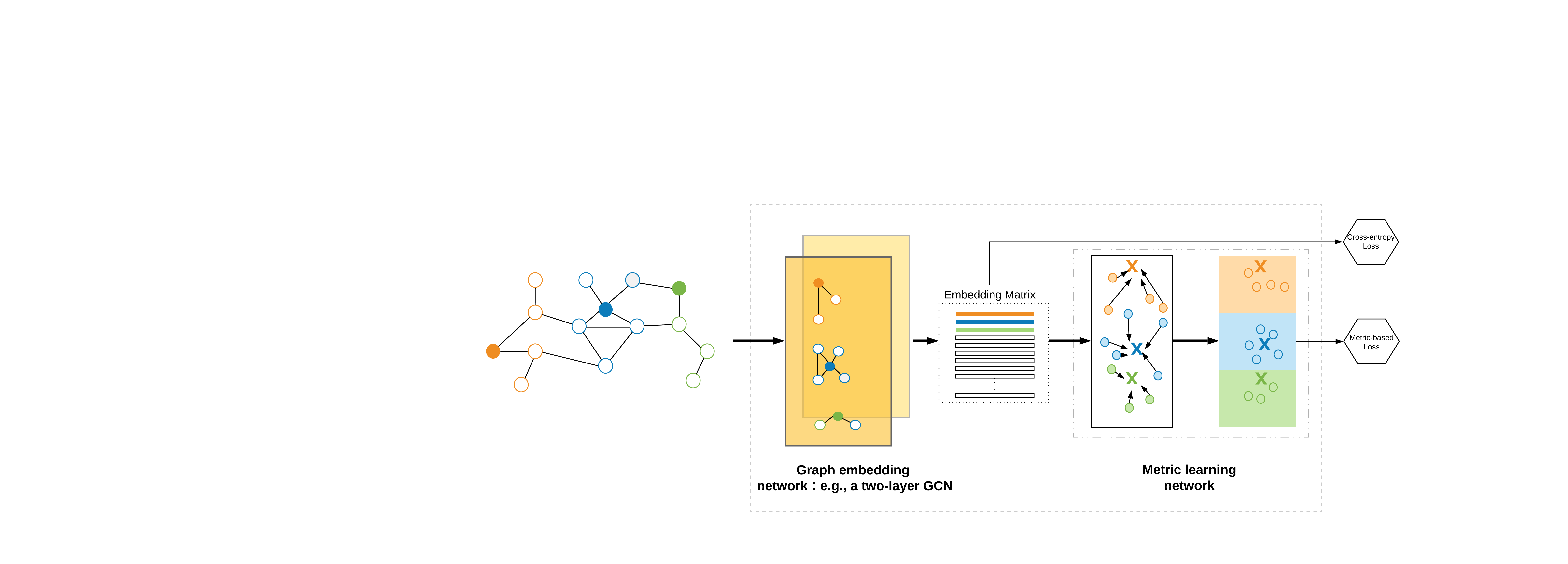}
  \caption{Illustration of the {\em Shoestring} framework: As an example, the input graph contains three types of node labels in three different colors, where the solid nodes are labeled and the rest are unlabeled. (1) A graph embedding network (a typical graph-based semi-supervised learning module, e.g., GCN) to learn a non-linear mapping of each node into an embedding vector; (2) A metric learning network to learn pair-wise similarity between each node representation and the centroid representation of each class in the low-dimensional semantic space. To optimize {\em Shoestring}, the full pipeline in our architecture is used.}
  \label{fig:framework}
  \end{center}
\end{figure*}


\section{Proposed Framework}
\label{sec:method}

In this section, we introduce our framework, called {\em Shoestring}, to address the problem of graph-based semi-supervised learning in the presence of severely limited labeled samples. The architecture of Shoestring is illustrated in Fig.~\ref{fig:framework}, which composed of two modules: a typical graph-based semi-supervised learning module/a graph embedding network to learn a non-linear mapping of each instance into an embedding vector (e.g., a two-layer GCN or label propagation module), and a metric learning module that learns the semantic similarity between each node representation and the centroid representation of each class in the low-dimensional semantic space. For simplicity, we first take semi-supervised GCNs as our prototypical model to illustrate our framework. We shall also discuss how to fit other semi-supervised learning methods into our framework, such as recent variations with graph filtering, IGCN and GLP.

Before we present our proposed framework, we first introduce the {\em manifold assumption} and {\em cluster assumption}~\cite{chapelle2009semi}, which are different from the smoothness assumption, but form the basis of our construction. The {\em manifold assumption} forms the basis of several semi-supervised learning methods in the literature, which indicates that the high-dimensional data lie on a low-dimensional manifold. The {\em cluster assumption} is one of the earliest forms of semi-supervised learning, which implies if data points/nodes are in the same cluster, they are likely to be of the same class.

As we discussed previously, the design basis of graph convolutional networks is the smoothness assumption (Laplacian smoothing). Its superior performance on semi-supervised classification tasks with sufficient labeled samples can also be interpreted as follows. The two-layer convolutional transformation tends to encourage the graph representations to lie on a low-dimensional manifold, such that the nodes can be classified distinctly in the embedding space. Motivated by this intuition, we seek to exploit the intrinsic structure of the data distribution in the embedding space, while the semi-supervised classification task is performed under fairly limited numbers of labeled samples. 

More specifically, the first component of {\em Shoestring} is a classical graph-based semi-supervised learning module, a two-layer GCN in our prototypical example, which is able to inject the graph structure into data representations by convolutional operations. With this transformation, the graph representations of the data are encouraged to lie on a low-dimensional manifold. In addition, we exploit a metric learning network that is able to learn a semantic metric space to represent the semantic similarity between the sparsely labeled and large numbers of unlabeled samples. Label assignments, for the unlabeled samples, are performed through transferring the semantic knowledge of the labeled samples. 

\begin{figure}[!t]
\begin{center}
    \begin{subfigure}[b]{0.25\textwidth}
  	\centering
    \includegraphics[width=\textwidth]{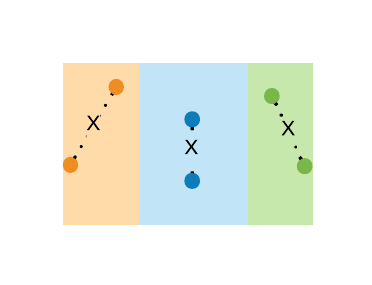}
    \caption{Per class centroid.}
    \label{fig:centroid}
  \end{subfigure}%
  \begin{subfigure}[b]{0.25\textwidth}
  \centering
    \includegraphics[width=\textwidth]{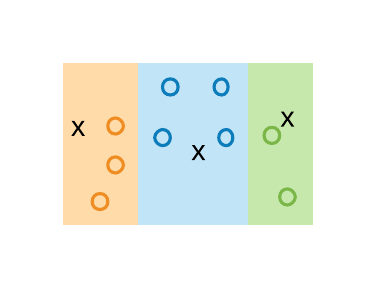}
    \caption{Label assignment.}
    \label{fig:clustering}
  \end{subfigure}
  \end{center}
 \label{fig:centroid_examples}
 \caption{Our metric learning network for semantic similarity learning and label assignments: (a) Three types of solid nodes are labeled samples in three classes, respectively. The class centroid of each class is calculated by the element-wise mean of the labeled samples in each class, as shown in cross signs. (b) Each unlabeled sample is assigned to the label of its nearest class centroid in the learned metric space.}
\end{figure}

Our metric learning network includes a similarity network to learn the semantic similarity between each node representation and the centroid representation of each class (the colored cross sign in Fig.~\ref{fig:framework}). In particular, the per-class centroid is the element-wise mean of its labeled samples in the embedding space (the output of the graph embedding network), shown in Fig.~\ref{fig:centroid}:

\begin{equation}
\label{eq:class_centroid}
c_{y_k} = \frac{1}{ |\mathcal{V}_{k}|}\sum_{(x_l,y_l) \in \mathcal{V}_{k}} Z_{\Theta, x_l}
\end{equation}

where $\mathcal{V}_{k} \subset \mathcal{V}_{l}$ and $Z_{\Theta, x_l}$ is the embedding vector of node $x_l$. Therefore, the output of the metric learning module contains the similarity values of each node to each class. Followed with a softmax (the output layer), the label of each unlabeled sample can be assigned to the class with the highest similarity value (its nearest class centroid), shown in Fig.~\ref{fig:clustering}:

\begin{equation}
\label{eq:classification}
p_{\Theta}(i=k|x_l) = \frac{\exp{[\mathbf{sim}(z_{\Theta, x_l}, c_{y_k})]}}{\sum_{i=1}^{K}{\exp{[\mathbf{sim}(z_{\Theta,x_l}, c_{y_i})]}}}
\end{equation}

 where $\mathbf{sim}$ is the distance function for similarity measurement in the low-dimensional embedding space. The underlying design intuition is that in the embedding space, the graph representations tend to lie on a low-dimensional manifold, in which closely clustered node representations tend to be assigned similar labels (the ``cluster assumption''). The underlying similarity function can be cosine similarity, or negative square Euclidean distance, etc. We will show the choice of a similarity metric is critical, where the performance varies from different datasets as well as various label rates.

\textbf{Objective function of {\em Shoestring}.} To optimize {\em Shoestring}, the full pipeline in our architecture is used. There are two components in our objective function. (1) The typical graph-based semi-supervised learning loss. (2) The metric-based learning loss. More specifically, in the GCN module, the first term is the cross-entropy loss as defined in Eq.~(\ref{eq:gcnloss}), while the second term is the metric-based cross-entropy loss. 

Assume we have $|\mathcal{V}_k|$ labeled samples from each class in the target domain. We compute the centroid representation $c_{y_k}$ for each class, by taking the element-wise mean of the $|\mathcal{V}_k|$ labeled samples, in the embedding space (the output of the classical graph embedding network). Thus, we can have the similarity vector for each labeled sample, where the $k$th element is the similarity between this sample and the centroid of class $c_{y_k}$. Therefore, the metric-based loss\footnote{Here we use term ``metric-based loss'' to differentiate the metric-based cross-entropy loss from the cross-entropy defined in Eq.~(\ref{eq:gcnloss}).} can be formulated as:
\begin{equation}
\label{eq:clustering}
\mathcal{L}_{\mathbf{me}} =- \sum_{(x_l, y_l)\in \mathcal{V}_l}\log{\frac{\exp[\mathbf{sim}(z_{x_l}, c_{y_l})]}{\sum_{i=1}^{K}{\exp{[\mathbf{sim}(z_{x_l}, c_{y_i})]}}}}
\end{equation}

Formally, the objective function of {\em Shoestring} is defined as follows:

\begin{equation}
\label{eq:shoestring}
\mathcal{L}_{\mathbf{{Shoestring}}} = \mathcal{L}_{\mathbf{ce}} + \lambda \mathcal{L}_{\mathbf{me}}
\end{equation}
where $\lambda$ is to control the degree of metric-based learning loss. After the optimization, {\em Shoestring} uses the forward propagation through the graph embedding network and the metric learning network, followed by a softmax (the output layer) to obtain the final label assignment.

Our proposed framework is fairly general that can be used to further boost the classification performance of several graph-based semi-supervised learning methods, while the number of labeled samples are severely limited. In particular, to fit label propagation, the label-efficient variations with graph filtering (IGCN, GLP) into our framework, we can just simply replace the graph embedding network module with any of these methods. In the experimental section, we will show empirically that {\em Shoestring} can indeed dramatically improve the classification accuracy of these methods, especially when there are only a few labeled samples.

\begin{table}[!ht]
\caption{Statistics description of citation networks.}
\begin{center}
\begin{small}
\begin{sc}
\setlength{\tabcolsep}{0.25em}
\begin{tabular}{ c| c |c |c| c}
\toprule
Dataset &Nodes & Edges & Classes & Features\\ 
\midrule
Cora &$2,708$ & $5.429$ & $7$ & $1,433$\\
CiteSeer &$3,327$ & $4,732$ & $6$ & $3,703$\\
PubMed&$19,717$&$44,338$&$3$&$500$\\
Large Cora&$11,881$&$64,898$&$10$&$3,780$\\
\bottomrule
\end{tabular}
\end{sc}
\end{small}	
\end{center}
\label{tab:data}
\end{table}

\section{Experiments}
\label{sec:exp}

We evaluate and compare {\em Shoestring} with state-of-the-art methods on semi-supervised document classification in citation networks, as well as a few-shot learning task for image classification on two datasets, e.g., miniImageNet and tieredImageNet.

\subsection{Performance Evaluation on Citation Networks}

\textit{Datasets.} Following~\cite{li2018deeper,li2019label}, we select four citation networks: Cora~\cite{mccallum2000automating}, CiteSeer~\cite{giles1998citeseer}, PubMed~\cite{sen2008collective} and Large Cora. The statistics of these datasets are summarized in Table~\ref{tab:data}. More specifically, for each citation network, we test several scenarios, each of which the number of labeled samples per class varies from  $1\sim 5$. In particular, we also test our framework under $20$ labeled samples per class to evaluate the performance of {\em Shoestring} with sufficient labeled samples.

\begin{table*}[!t]
\caption{Document classification accuracy on citation networks ($\%$).}
\begin{adjustbox}{center}
\begin{small}
\setlength{\tabcolsep}{0.25em}
\begin{tabular}{ c l || c c c c | c c c c | c c c c }
\cmidrule[1pt]{2-14}
&Label Rate & \multicolumn{4}{{c}}{1 label per class}  & \multicolumn{4}{{c}}{2 labels per class} & \multicolumn{4}{{c}}{5 labels per class} \\
&Dataset&Cora&CiteSeer&PubMed&Large Cora&Cora&CiteSeer&PubMed&Large Cora&Cora
 &CiteSeer&PubMed&Large Cora\\
\cmidrule{2-14}
&LP&$43.6_{(0.1)}$&$30.6_{(0.1)}$&$49.8_{(0.2)}$&$24.3_{(0.3)}$&$53.1_{(0.1)}$&$33.0_{(0.1)}$&$56.1_{(0.2)}$&$37.2_{(0.3)}$&$60.6_{(0.1)}$&$41.5_{(0.1)}$&$64.5_{(0.2)}$&$42.1_{(0.3)}$\\
&GCN&$39.5_{(0.6)}$&$34.1_{(0.9)}$&$50.8_{(4.8)}$&$28.1_{(3.8)}$&$51.7_{(0.7)}$&$45.5_{(1.0)}$&$59.9_{(5.0)}$&$39.6_{(3.9)}$&$68.7_{(0.6)}$&$57.0_{(0.9)}$&$69.6_{(4.8)}$&$51.8_{(3.8)}$\\
&ST-CT&$54.7_{(5.3)}$&$48.5_{(8.4)}$&$59.3_{(51)}$&$31.8_{(36)}$&$62.7_{(5.5)}$&$51.3_{(8.4)}$&$67.3_{(51)}$&$41.6_{(35)}$&$73.1_{(5.6)}$&$63.5_{(8.8)}$&$71.0_{(53)}$&$53.4_{(36)}$\\
&IGCN\textsubscript{(RNM)}&$41.5_{(0.6)}$&$33.1_{(1.0)}$&$51.4_{(4.9)}$&$30.9_{(4.2)}$&$62.6_{(0.7)}$&$44.5_{(1.0)}$&$60.4_{(5.3)}$&$44.9_{(4.5)}$&$71.2_{(0.6)}$&$57.6_{(0.9)}$&$70.5_{(4.9)}$&$55.4_{(4.2)}$\\
&IGCN\textsubscript{(AR)}&$42.3_{(1.0)}$&$33.0_{(1.3)}$&$52.1_{(5.7)}$&$31.6_{(8.8)}$&$62.7_{(1.7)}$&$44.9_{(1.9)}$&$61.6_{(8.1)}$&$45.3_{(9.4)}$&$72.1_{(1.0)}$&$58.1_{(1.2)}$&$71.1_{(5.7)}$&$55.7_{(8.8)}$\\
&GLP\textsubscript{(RNM)}&$38.4_{(0.4)}$&$37.0_{(0.7)}$&$54.7_{(0.8)}$&$30.2_{(2.2)}$&$59.6_{(0.4)}$&$46.0_{(0.6)}$&$60.6_{(0.6)}$&$45.2_{(2.0)}$&$72.2_{(0.4)}$&$59.2_{(0.7)}$&$69.9_{(0.8)}$&$55.4_{(1.5)}$\\
&GLP\textsubscript{(AR)}&$37.7_{(4.0)}$&$37.4_{(19)}$&$55.8_{(9.1)}$&$27.8_{(26)}$&$57.7_{(3.4)}$&$46.1_{(16)}$&$61.7_{(7.6)}$&$44.8_{(26)}$&$71.1_{(3.9)}$&$59.4_{(19)}$&$71.2_{(9.0)}$&$55.7_{(13)}$\\
\cmidrule{2-14}

&GCN&$60.2_{(0.9)}$&$52.2_{(1.3)}$&$60.3_{(6.1)}$&$48.0_{(4.0)}$&$68.3_{(0.9)}$&$57.7_{(1.3)}$&$63.5_{(5.7)}$&$52.8_{(4.1)}$&$73.0_{(1.2)}$&$64.2_{(1.5)}$&$68.6_{(6.3)}$&$58.9_{(4.5)}$\\
&IGCN\textsubscript{(RNM)}&$69.1_{(1.0)}$&$\textbf{57.9}_{(1.4)}$&$63.3_{(6.2)}$&$\textbf{54.6}_{(4.4)}$&$73.0_{(1.0)}$&$\textbf{61.7}_{(1.4)}$&$64.9_{(6.2)}$&$57.3_{(4.5)}$&$76.4_{(1.3)}$&$\textbf{65.8}_{(1.6)}$&$69.0_{(7.1)}$&$61.4_{(5.1)}$\\
&IGCN\textsubscript{(AR)}&$\textbf{70.1}_{(2.4)}$&$\textbf{58.3}_{(2.7)}$&$64.7_{(11)}$&$\textbf{56.0}_{(8.3)}$&$\textbf{73.3}_{(2.4)}$&$\textbf{61.9}_{(2.7)}$&$66.4_{(11)}$&$\textbf{58.1}_{(8.5)}$&$\textbf{76.5}_{(3.0)}$&$\textbf{65.9}_{(3.4)}$&$70.0_{(13)}$&$61.6_{(9.5)}$\\
&GLP\textsubscript{(RNM)}&$69.3_{(0.6)}$&$57.6_{(0.8)}$&$63.3_{(0.8)}$&$54.2_{(2.2)}$&$72.8_{(0.6)}$&$61.3_{(0.8)}$&$65.0_{(0.8)}$&$56.4_{(2.7)}$&$75.7_{(0.8)}$&$65.0_{(1.1)}$&$67.9_{(1.3)}$&$59.9_{(3.3)}$\\
\rot{\rlap{~{\em Shoestring}-COS}}
&GLP\textsubscript{(AR)}&$\textbf{69.8}_{(3.7)}$&$\textbf{58.1}_{(17)}$&$\textbf{65.2}_{(7.7)}$&$\textbf{55.5}_{(26)}$&$\textbf{73.5}_{(3.7)}$&$\textbf{61.7}_{(17)}$&$66.2_{(7.6)}$&$\textbf{57.7}_{(26)}$&$76.3_{(4.9)}$&$\textbf{65.4}_{(21)}$&$69.7_{(11)}$&$61.5_{(32)}$\\
\cmidrule{2-14}

&GCN&$60.7_{(1.3)}$&$51.0_{(1.5)}$&$62.1_{(6.1)}$&$46.5_{(4.7)}$&$67.4_{(1.2)}$&$55.5_{(1.5)}$&$64.6_{(6.1)}$&$53.9_{(4.7)}$&$74.2_{(1.3)}$&$62.2_{(1.5)}$&$71.4_{(6.0)}$&$\textbf{62.0}_{(4.7)}$\\
&IGCN\textsubscript{(RNM)}&$69.6_{(1.4)}$&$54.5_{(1.7)}$&$64.4_{(6.7)}$&$53.3_{(5.2)}$&$73.1_{(1.4)}$&$58.6_{(1.7)}$&$\textbf{67.1}_{(6.7)}$&$\textbf{57.7}_{(5.1)}$&$76.4_{(1.4)}$&$63.8_{(1.7)}$&$71.7_{(6.8)}$&$\textbf{62.0}_{(5.1)}$\\
&IGCN\textsubscript{(AR)}&$\textbf{70.1}_{(2.8)}$&$54.9_{(3.2)}$&$\textbf{66.4}_{(12)}$&$53.2_{(9.2)}$&$\textbf{73.4}_{(2.8)}$&$59.3_{(3.2)}$&$\textbf{67.3}_{(12)}$&$57.6_{(9.0)}$&$\textbf{76.7}_{(2.8)}$&$64.3_{(3.2)}$&$\textbf{73.1}_{(12)}$&$61.7_{(9.1)}$\\
&GLP\textsubscript{(RNM)}&$68.1_{(0.9)}$&$52.3_{(1.1)}$&$64.1_{(1.2)}$&$49.7_{(2.8)}$&$72.3_{(0.9)}$&$57.3_{(1.1)}$&$65.5_{(1.2)}$&$56.4_{(2.8)}$&$75.8_{(0.9)}$&$62.5_{(1.1)}$&$\textbf{72.1}_{(1.1)}$&$61.3_{(2.8)}$\\
\rot{\rlap{~{\em Shoestring}-L2}}
&GLP\textsubscript{(AR)}&$68.0_{(4.2)}$&$53.5_{(17)}$&$\textbf{65.5}_{(8.3)}$&$49.1_{(27)}$&$72.9_{(4.1)}$&$57.9_{(17)}$&$\textbf{66.5}_{(8.2)}$&$56.9_{(26)}$&$\textbf{76.7}_{(4.1)}$&$63.3_{(17)}$&$\textbf{74.0}_{(8.2)}$&$\textbf{63.1}_{(26)}$\\
\cmidrule[1pt]{2-14}

\end{tabular}
\end{small} 
\end{adjustbox}
\label{tab:global_ranking}
\end{table*} 

\noindent\textit{Baselines.} As {\em Shoestring} aims for boosting the learning performance of graph-based semi-supervised learning methods, we implemented several existing models as the base models of {\em Shoestring} and compared with their original implementations. These methods are GCN~\cite{kipf2016semi}, IGCN\textsubscript{(RNM)}, IGCN\textsubscript{(AR)}, GLP\textsubscript{(RNM)}, and GLP\textsubscript{(AR)}. In addition, we also compared with the methods that training GCN with self-training and co-training~\cite{li2018deeper} (For simplicity, we call this set of methods {\em ST-CT}). Experimental results are averaged over $20$ runs to ensure statistical significance.

It is worth mentioning that IGCN and GLP~\cite{li2019label} are the state-of-the-art methods for semi-supervised learning under limited labeled samples. They are variations of GCN and LP~\cite{zhou2004learning} from the perspective of graph filtering. More specifically, IGCN\textsubscript{(RNM)} and IGCN\textsubscript{(AR)} change the renormalization of the adjacency matrix of the original GCN to Auto-Regressive filter (AR) and renormalization filter (RNM), respectively. GLP\textsubscript{(RNM)} and GLP\textsubscript{(AR)} propagate node features through the graph instead of propagating labels in LP. The input node features are filtered using Auto-Regressive filter (AR) or renormalization filter (RNM) for GLP\textsubscript{(AR)} and GLP\textsubscript{(RNM)} respectively. A classifier is trained on propagated features to generate the labels.

For ST-CT~\cite{li2018deeper}, there are four different proposals, including co-training, self-training, union, and intersection to train GCN so as to improve the learning performance. More specifically, co-training is a GCN with a random walk model that can add the nearest neighbors of the labeled nodes to expand the labeled set iteratively. Self-training is an iterative process, where a classifier assigns the labels for the unlabeled samples which have been classified with confidence in the previous step. Union expands the training set with both random walk and GCN. Intersection, similar to union, also uses two methods but only uses the predictions that are in common. Due to the space limitation, we reported the best accuracy among these four methods.

\noindent{\em Similarity metrics.} In our similarity network, we used three types of similarity metrics: distance-based similarity according to L1 and L2 respectively (negative distance value as the similarity), and cosine similarity. More specifically, L1 calculates the distance between two nodes by adding the absolute differences of their feature embeddings, while L2 adds the squares differences of the feature embeddings. Cosine similarity (COS), on the other hand, is a similarity measurement between two non-zero vectors of an inner product space. 

All the experiments were performed on a machine with Intel Core i7-9700K $8$-core $3.6$GHz CPU, $32$GB RAM, $500$GB SSD, and GeForce GTX 1660 Ti GPU.

\noindent\textit{Parameter settings.} For LP, GCN, IGCN, and GLP, we use the same setting as in ~\cite{li2019label}: $0.01$ learning rate, $0.5$ dropout rate, $5*10^{-4}$ weight decay, $200$ epochs, $16$ hidden units for Cora, CiteSeer, PubMed and $64$ hidden units for Large Cora. The weight of the metric-based cross-entropy loss is tuned amongst $\{0.001, 0.01, 0.05, 0.1\}$ and is set to $0.01, 0.05, 0.001$ for COS, L1 and L2 similarity metrics respectively.

\noindent\textit{Results analysis.} The results for  $1$, $2$, and $5$ labeled samples are reported in Table~\ref{tab:global_ranking}. We highlighted the top-$3$ classification accuracies in bold. Due to the space limitation, we reported the results for $3$ and $4$ labeled samples and all of the results using L1 distance metric in the Appendix. A first conclusion that we can draw from these experiments is that no similarity metric is uniformly better than the others. We can also observe that IGCN\textsubscript{(AR)} and GLP\textsubscript{(AR)} under {\em Shoestring} with cosine similarity perform the best in overall cases. In particular, for $5$ labeled samples per class, there is a $\sim5\%$ improvement with our proposed framework as compared to the original implementation. As the label rates get smaller, the improvement increases significantly, up to $32.1\%$ performance gain on Cora for GLP\textsubscript{(AR)} with $1$ labeled sample per class, which shows the label-efficiency of our methods. 

To further investigate the performance of our model on datasets with a larger portion of labeled samples, we test the scenario with $20$ labeled samples per class. The results are shown in Table~\ref{tab:20l} with the best accuracy highlighted in bold. One of the interesting results of our framework is when the labeled samples are sufficiently large enough. Indeed, it has been shown that semi-supervised learning methods under our {\em Shoestring} can be very useful and the results from Table~\ref{tab:20l} exhibit better classification performance over the baseline methods. There could be a possible explanation on this fact that these semi-supervised learning models have already been effective and reliable to generate smooth and representative features for subsequent classification, when the number of labeled samples is significantly large. Augmented with a metric learning network, which is designed on the basis of the manifold assumption and cluster assumption in the embedding space, it can achieve a further performance gain, up to $1.7\%$ on Large Cora.

\begin{figure}[!ht]
\begin{center}
  \begin{subfigure}[b]{0.23\textwidth}
    \includegraphics[width=\textwidth]{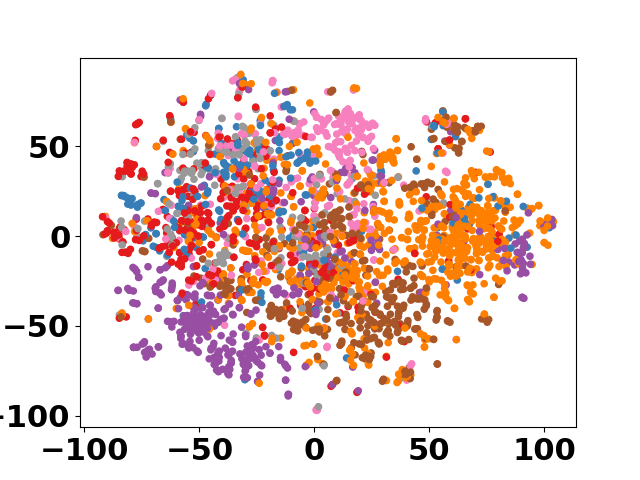}
    \caption{Raw representations of Cora}
    \label{fig:raw}
  \end{subfigure}
  \hfill
  \begin{subfigure}[b]{0.23\textwidth}
    \includegraphics[width=\textwidth]{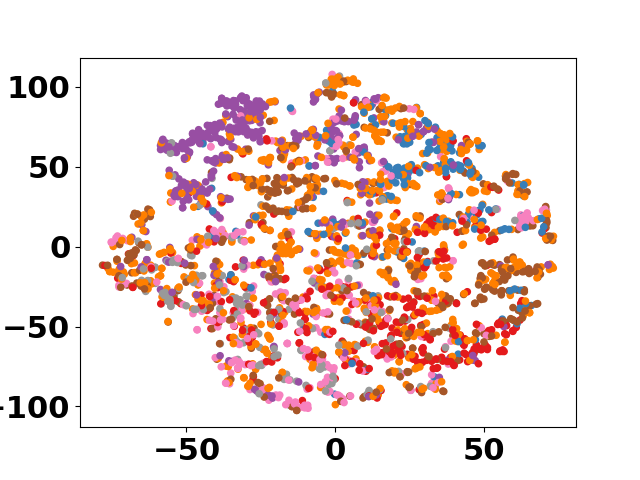}
    \caption{GCN with 1 label per class}
    \label{fig:gcn}
  \end{subfigure}
  \begin{subfigure}[b]{0.23\textwidth}
    \includegraphics[width=\textwidth]{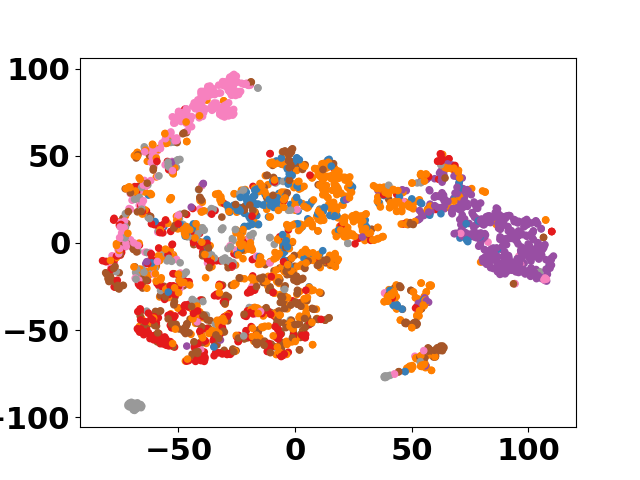}
    \caption{{\em Shoestring}-COS with 1 label per class}
    \label{fig:1-cos}
  \end{subfigure}
  \hfill
  \begin{subfigure}[b]{0.23\textwidth}
    \includegraphics[width=\textwidth]{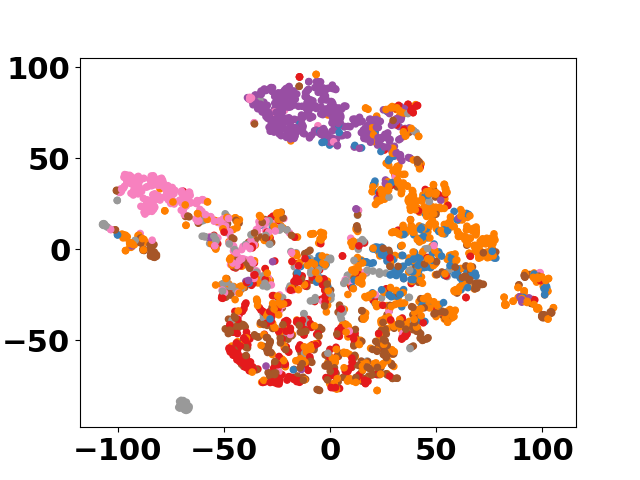}
    \caption{{\em Shoestring}-L2 with 1 label per class}
    \label{fig:1-l2}
  \end{subfigure}
  \begin{subfigure}[b]{0.23\textwidth}
    \includegraphics[width=\textwidth]{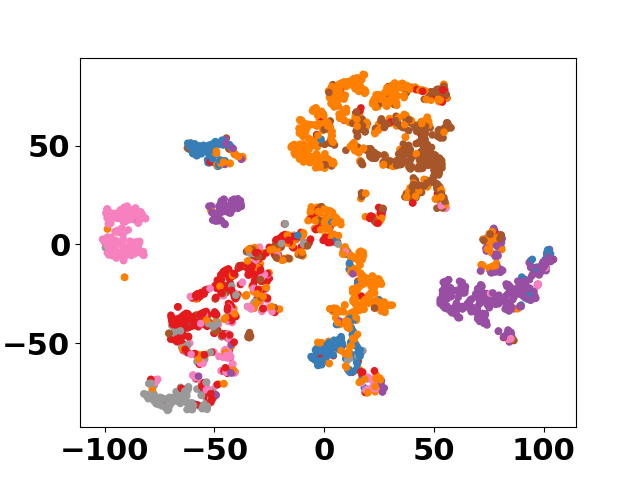}
    \caption{{\em Shoestring}-COS with 5 labels per class}
    \label{fig:5-cos}
  \end{subfigure}
  \hfill
  \begin{subfigure}[b]{0.23\textwidth}
    \includegraphics[width=\textwidth]{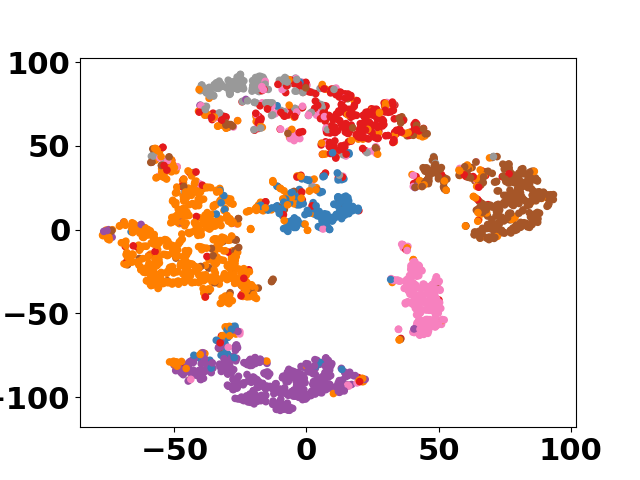}
    \caption{{\em Shoestring}-COS with 20 labels per class}
    \label{fig:20-cos}
   \end{subfigure}
 \end{center}
 
 \caption{Visualization of Cora features.}
\label{fig:t_sne}
\end{figure}

The reason for high performance even with severely limited labeled samples is that, {\em Shoestring} can locate the centroid for each class and generate labels based on the cluster assumption and manifold assumption, which enables transferring as much knowledge as possible from sparsely labeled samples to a large number of unlabeled samples in the embedding space. To clearly visualize the improvement, Fig.~\ref{fig:t_sne} shows the raw features of Cora, its feature embeddings learned with one labeled sample per class based on the original GCN~\cite{kipf2016semi}, and the feature embeddings learned based on {\em Shoestring}-COS and {\em Shoestring}-L2, respectively. The results show that GCN performs poorly with one labeled sample per class, while our proposed framework can cluster more compactly, as shown in Fig.~\ref{fig:1-cos} and Fig.~\ref{fig:1-l2}. The feature embeddings learned with $5$ labels and $20$ labels with {\em Shoestring}-COS are also shown in Fig.~\ref{fig:5-cos} and Fig.~\ref{fig:20-cos}, respectively.

\noindent{\textbf{Computation cost.}} The time needed for computing centroid and relative distance for similarity measurement in each iteration is corresponding with the number of classes. As the benchmarking datasets do not have a significant amount of classes, the time efficiency of {\em Shoestring} is comparable with the original implementations. As reported in Table~\ref{tab:global_ranking}, the numbers in brackets are the computation time of each model to perform classification. For $1$-labeled sample per class, there is only a $0.5$ second increase in time on average with a $20\%$ performance gains on average.

\begin{table}[!t]
\caption{Document classification accuracy on citation networks with $20$ labeled samples ($\%$).}
\begin{center}
\begin{small}
\setlength{\tabcolsep}{0.25em}
\begin{tabular}{ l || c c c c}
\cmidrule[1pt]{1-5}
Label Rate & \multicolumn{4}{{c}}{20 labels per class}\\
Dataset&Cora&CiteSeer&PubMed&Large Cora\\
\cmidrule{1-5}
LP&$67.8_{(0.1)}$&$47.7_{(0.1)}$&$73.3_{(0.2)}$&$52.5_{(0.3)}$\\
GCN&$79.8_{(0.7)}$&$68.1_{(0.9)}$&$78.0_{(5.4)}$&$67.4_{(3.8)}$\\
ST-CT&$80.1_{(5.8)}$&$\textbf{70.1}_{(9.1)}$&$77.6_{(54)}$&$66.0_{(18)}$\\
IGCN\textsubscript{(RNM)}&$80.9_{(0.7)}$&$68.4_{(1.0)}$&$77.6_{(5.5)}$&$68.5_{(3.9)}$\\
IGCN\textsubscript{(AR)}&$81.3_{(1.4)}$&$68.6_{(1.7)}$&$78.5_{(8.5)}$&$68.8_{(6.4)}$\\
GLP\textsubscript{(RNM)}&$80.7_{(0.3)}$&$67.7_{(0.4)}$&$77.7_{(0.4)}$&$68.1_{(1.5)}$\\
GLP\textsubscript{(AR)}&$81.2_{(1.8)}$&$68.4_{(8.6)}$&$78.8_{(3.9)}$&$68.7_{(13)}$\\
\cmidrule{1-5}
Ours&$\textbf{81.9}_{(2.1)}$&$69.5_{(2.4)}$&$\textbf{79.7}_{(4.4)}$&$\textbf{70.5}_{(4.7)}$\\
\cmidrule[1pt]{1-5}
\end{tabular}
\end{small} 
\end{center}
\label{tab:20l}
\end{table}

\subsection{Performance Evaluation on Few-Shot Image Classification}

Our proposed framework can also be used for few-shot image classification. Few-shot learning~\cite{finn2017model} is to learn a classifier that generalizes well even when trained with a limited number of training instances per class. An episodic meta-learning strategy~\cite{vinyals2016matching}, due to its generalization performance, has been adopted by many works on few-shot learning. To achieve lager improvements with limited numbers of training instances, several previous works proposed to consider the relationships between instances and representing the data into a graph~\cite{garcia2017few,liu2018learning}. In particular, TPN~\cite{liu2018learning} proposed to propagate labels between data instances for unseen classes via episodic meta-learning. Here, we replace the label propagation module with {\em Shoestring} in each episode training of TPN and test its performance on the few-shot image classification task.

\noindent\textit{Datasets.} For fair comparisons with previous works, we use two datasets, miniImageNet and tieredImageNet, and follow the data preprocessing and split from~\cite{liu2018learning}. The miniImageNet dataset is a subset of ImageNet dataset and designed for few-shot classification. It has $100$ classes with $64$ classes for training, $16$ for validation, $20$ for test, and $600$ examples per class. Similar to miniImageNet, tieredImageNet is also a subset of ImageNet dataset, which has $608$ classes and the average number of examples for each class is $1,281$. It has a hierarchical structure with $34$ categories which are separated to $20$ for training, $6$ for validation, and $8$ for test. 

\noindent\textit{Baselines.} Except for TPN~\cite{liu2018learning}, we also compared with the state-of-the-art method, MetaOptNet~\cite{lee2019meta}. This model adapted the meta-learning framework with different convex base learners for few-shot learning. In particular, the framework was incorporated with ridge regression and support vector machines, called MetaOptNet-RR and MetaOptNet-SVM, respectively. For fair comparisons, both TPN and MetaOptNet used the standard $4$-layer convolutional network with $64$ filters per layer as their feature embedding architecture.
 
\noindent\textit{Parameter settings.} Our implementation followed the parameter settings in~\cite{liu2018learning}, where the hyper-parameter $k$ of the k-nearest neighbor graph is set to $20$, label propagation parameter $\alpha$ is set to $0.99$, the query number is $15$, and the results are averaged over $600$ randomly generated episodes from the test set. In addition, the learning rate is set to $10^{-3}$ initially and then is halved every $10,000$ episodes for miniImageNet and $25,000$ episodes for tieredImageNet, respectively. The tests are conducted under the semi-supervised condition with $5$-way $1$-shot and $5$-way $5$-shot for both datasets. 

\noindent\textit{Results analysis.} The results are shown in Table~\ref{tab:TPN_semi} with the top accuracy of each category highlighted in bold. The results of benchmarking datasets are directly obtained from their papers. From experiments, we observe that the cosine similarity is best suited for image classification and, therefore, we only include results from this method. {\em Shoestring}-TPN\textsubscript{(COS)} outperformed all baseline methods. In particular, {\em Shoestring}-TPN\textsubscript{(COS)} achieved significant gains on miniImageNet ($2.57\%\sim3.59\%$) and tieredImageNet ($1.05\%\sim2.70\%$), respectively. In addition, TPN under {\em Shoestring} leads to state-of-the-art performance as compared to MetaOptNet, demonstrating the effectiveness of {\em Shoestring} on few-shot image classification tasks. We can observe that the improvement for $1$-shot learning is even higher than that of $5$-shot, $1.765\%$ and $1.05\%$ respectively, showing that {\em Shoestring} can provide with more superior performance in severely limited labeled samples. 

\begin{table}[!t]
\caption{Classification accuracy ($\%$) on few-shot image classification on miniImageNet and tieredImageNet ($5$-way).}
\begin{center}
\begin{small}
\setlength{\tabcolsep}{0.25em}
\begin{tabular}{ c || c c | c c }
\toprule
& \multicolumn{2}{{c}}{miniImageNet}&\multicolumn{2}{{c}}{tieredImageNet}\\
Model&$1$-shot&$5$-shot&$1$-shot&$5$-shot\\
\midrule
TPN&$52.78$&$66.42$&$55.74$&$71.01$\\
MetaOptNet-RR&$53.23$&$69.51$&$54.63$&$72.11$\\
MetaOptNet-SVM&$52.87$&$68.76$&$54.71$&$71.79$\\
\midrule
{\em Shoestring}-TPN\textsubscript{(COS)}&$\textbf{55.35}$&$\textbf{70.01}$&$\textbf{56.79}$&$\textbf{73.71}$\\
\bottomrule
\end{tabular}
\end{small} 
\end{center}
\label{tab:TPN_semi}
\end{table}

\section{Concluding Remarks}
\label{sec:conclusion}

In this paper, we advanced the graph-based semi-supervised learning paradigm towards a scenario where labeled data are severely limited. We proposed a new framework, called {\em Shoestring}, which is designed on the basis of the manifold assumption and cluster assumption in the embedding space. The experiments for both document classification on citation networks and few-shot learning image classification show strong benefits of using {\em Shoestring}, resulting in new state-of-the-art results across overall cases. The key factor that determines the performance of our proposed framework is that, with the metric learning network, {\em Shoestring} can transfer the semantic knowledge of a limited number of labeled samples to a large number of unlabeled samples. Therefore, even with just a few labeled samples, {\em Shoestring} can outperform all of the baseline methods. We empirically show the choice of similarity metrics in our framework is critical. One strategy to fit different datasets with different similarity metrics is to learn an adaptive similarity function. We leave this as our future work.

{\small
\bibliographystyle{ieee_fullname}
\bibliography{main}
}

\end{document}